\pdfoutput=1

\documentclass[11pt]{article}
\usepackage[dvipsnames]{xcolor}
\usepackage[]{EMNLP2022}

\usepackage{times}
\usepackage{latexsym}
\usepackage{graphicx}
\usepackage{amsmath}
\usepackage{listings}
\usepackage{multirow}
\usepackage{url}

\usepackage[T1]{fontenc}
\usepackage{ulem}

\usepackage[utf8]{inputenc}

\usepackage{microtype}

\usepackage{inconsolata}

\title{Fast Vocabulary Transfer for Language Model Compression}

\author{Leonidas Gee\\
  Expert.ai, Italy \\
  \texttt{lgee@expert.ai} \\ \And
  Andrea Zugarini\\
  Expert.ai, Italy \\
  \texttt{azugarini@expert.ai} \\ \AND
  Leonardo Rigutini\\
  Expert.ai, Italy \\ and \\ University of Siena \\
  \texttt{lrigutini@expert.ai} \\
  \\ \And
  Paolo Torroni \\
  Department of Computer Science and \\ Engineering, \\ University of Bologna, Italy\\
  \texttt{paolo.torroni@unibo.it} \\
}

\def\Bs{\mbox{\boldmath $s$}}

\def\Bt{\mbox{\boldmath $t$}}

\def\tok{\mathcal{T}}
\def\tokgen{\mathcal{T}_{gen}}
\def\domgen{\mathcal{D}_{gen}}
\def\tokin{\mathcal{T}_{in}}
\def\domin{\mathcal{D}_{in}}

\begin{document}
\maketitle

\begin{abstract}

Real-world business applications require a trade-off between language model performance and size. We propose a new method for model compression that relies on vocabulary transfer. We evaluate the method on various vertical domains and downstream tasks. Our results indicate that vocabulary transfer can be effectively used in combination with other compression techniques, yielding a significant reduction in model size and inference time while marginally compromising on performance.




\end{abstract}

\section{Introduction}
In the last few years, many NLP applications have been relying more and more on large pre-trained Language Models (LM)~\cite{bert,roberta,he2020deberta}. Because larger LMs, on average, exhibit higher accuracy, a common trend has been to increase the model's size. 
Some LMs like GPT-3 \cite{brown2020language} and BLOOM\footnote{\url{https://bigscience.huggingface.co/blog/bloom}} have reached hundreds of billion parameters. 
However, these models' superior performance comes at the cost of a steep increase in computational footprint, both for development and for inference, 
ultimately hampering their adoption in real-world business use-cases.
%
Besides models that only a few hi-tech giants can afford, like GPT-3, 
even smaller LMs with hundreds of million parameters could be too expensive or infeasible for certain products. 
For one thing, despite being tremendously cheaper than their bigger cousins, fine-tuning, deploying and maintaining large numbers of such models (one for each downstream task) soon becomes too expensive. Furthermore, latency and/or hardware requirements may limit their applicability to specific use-cases. 
For all these reasons, significant efforts -- in both academic and industry-driven research -- are oriented towards the designing of solutions to drastically reduce the costs of LMs.

\begin{figure*}[ht]
  \centering
  \includegraphics[scale=0.75]{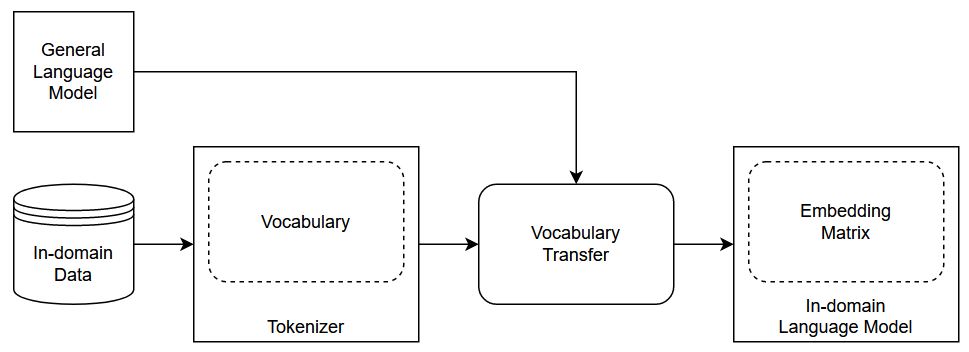}
  \caption{Sketch of the VT procedure. First, the vocabulary is constructed on the in-domain data, then an embedding is assigned to each token, transferring information from the pre-trained representations of the general-purpose language model.}
\end{figure*}

Recently, several attempts have been made to make these models smaller, faster and cheaper, while retaining most of their original performance \cite{gupta2015deep,quantization}. 
Notably, Knowledge Distillation (KD) \cite{hinton2015distilling} is 
a teacher-student framework, whereby the teacher consists of a pre-trained large model and the student of a smaller one. The teacher-student framework requires that both the teacher and the student estimate the same probability distribution. While the outcome is a smaller model, yet, this procedure constrains the student to operate with the same vocabulary as the teacher in the context of Language Modeling. 

In this work, we explore a method for further reducing an LM’s size by compressing its vocabulary through the training of a tokenizer in the downstream task domain.
The tokenizer~\cite{sennrich2016neural,schuster2012japanese,kudo2018sentencepiece} is a crucial part of modern LMs. In particular, moving from word to subword-level, the tokenization solves two problems: vocabulary explosion and unknown words. Moreover, the capability to tokenize text effectively in any domain is key for the massive adoption of pre-trained general-purpose LMs 
fine-tuned on downstream tasks. Indeed, tokenizers are still able to process out-of-distribution texts at the cost of producing frequent word splits into multiple tokens.

However, the language varies significantly in vertical domains or, more generally, in different topics. 
Hence, ad-hoc tokenizers, trained on the domain statistics, may perform a more efficient tokenization, reducing on average the length of the tokenized sequences. This is important since compact and meaningful inputs could reduce computational costs, while improving performance. 
Indeed, memory and time complexity of attention layers grows quadratically with respect to the sequence length~\cite{vaswani2017attention}. 
Furthermore, a vertical tokenizer may require a smaller vocabulary, which also affects the size of the embedding matrix, hence further reducing the model's size.

Following this intuition, we propose a Vocabulary Transfer (VT) technique to adapt LMs to in-domain, smaller tokenizers, in order to further compress and accelerate them. This technique is complementary to the aforementioned model compression methods and independent of the type of tokenizer. As a matter of fact, we apply it in combination with KD. 

Our experiments show that VT achieves an inference speed-up between x1.07 and x1.40, depending on the downstream task, with a limited  performance drop, and that a combination of VT with KD yields an overall reduction up to x2.76. 

The paper is organized as follows. After reviewing related works in Section~\ref{sec:relworks}, we present the methodology in Section~\ref{sec:methodology}, we then outline the experiments in Section~\ref{sec:experiments} and draw our conclusions in Section~\ref{sec:conclusions}. 

\section{Related Works}\label{sec:relworks}

The goal of Model Compression is to shrink and optimize neural architectures, while retaining most of their initial performance. Research on LM compression has been carried out following a variety of approaches like quantization~\cite{gupta2015deep,quantization}, pruning~\cite{prunenoprune,sixteenheads} knowledge distillation~\cite{distilbert,jiao2020tinybert,wang2020minilm}, and combinations thereof~\cite{polino2018model}.

A most popular distillation approach in NLP was proposed by \citet{distilbert}. The obtained model, called DistilBERT, is a smaller version of BERT, with the same architecture but half the layers,  trained to imitate the full output distribution of the teacher (a pre-trained BERT model). DistilBERT has a 40\% smaller size than BERT and retains 97\% of its language understanding capabilities. This enables a 60\% inference-time speedup. Further compression was achieved by~\citet{jiao2020tinybert} by adding transformer-layer, prediction-layer and embedding-layer distillation. The resulting model, TinyBERT, is 10 times smaller than BERT, with only four layers and reduced embeddings sizes. Related methods were proposed \cite{sun2020mobilebert,wang2020minilm}, achieving similar compression rates.
All these works focus on the distillation of general-purpose language models.  \citet{gordon-duh-2020-distill}  investigated the interaction between KD and Domain Adaptation.

Little focus has been devoted thus far to the role of tokenization in the context of model compression. Even in domain adaptation \cite{gordon-duh-2020-distill}, the vocabulary was kept the same. Both the versatility of the subword-level tokenization, and the constraints imposed by the teacher-student framework (same output distribution), discouraged such investigations. Recently, \citet{samenko2021fine} presented an approach for transferring the vocabulary of an LM into a new vocabulary learned from new domain, with the purpose of boosting the performance of the fine-tuned model. To the best of our knowledge, we are the first to study VT in the scope of model compression.





\section{Vocabulary Transfer}
\label{sec:methodology}
%
Let us consider a LM, trained on a general-purpose domain $\domgen$ and associated with a vocabulary ${\cal V}_{gen}$. Such a vocabulary is used by the LM's tokenizer in order to produce an encoding of the input string via an embedding matrix $E_{gen}$ defined on ${\cal V}_{gen}$. More specifically, a tokenizer is a function that maps a textual string into a sequence of symbols of a given vocabulary ${\cal V}$. Let $\tok$ be a tokenizer associated with a vocabulary ${\cal V}$ and a string $\Bs$, we have  $\tok: \Bs \rightarrow (\Bt_1, \ldots, \Bt_n), \Bt_i \in {\cal V},  \forall i=1,\ldots,n$. Hence, the vocabulary of the tokenizer determines how words in a text are split, whether as words, subwords, or even characters. These symbols, which define the LM's vocabulary, are statistically determined by training the tokenizer to learn the distribution of a dataset.

Now, let us consider a vertical domain $\domin$, also referred as \textit{in-domain}. For the reasons discussed earlier, a vocabulary ${\cal V}_{in}$ specialized on $\domin$ itself better fits the language distribution than ${\cal V}_{gen}$. Unfortunately, with a new vocabulary, embedding representations associated with the tokens of  ${\cal V}_{gen}$ would be lost. Thus, VT aims to initialize ${\cal V}_{in}$ by re-using most of the information learned from the LM pre-trained on $\domgen$.
 Once the new tokenizer $\tokin$ has been trained on the in-domain dataset $\domin$ using a given vocabulary size,
$\tokin$ will be different from the LM's tokenizer $\tokgen$. However, the two tokenizers' vocabularies ${\cal V}_{gen}$ and ${\cal V}_{in}$ may still have a large portion of their symbols in common.
Our objective is to transfer most of the information from ${\cal V}_{gen}$ into ${\cal V}_{in}$. To this end, we first define a mapping between each symbol in ${\cal V}_{in}$ and a set of symbols in ${\cal V}_{gen}$. Then, we define an assignment criterion, based on the mapping, to obtain the embeddings for the tokens of $\tokin$.

One such criterion, called Vocabulary Initialization with Partial Inheritance (VIPI), was defined by~\citet{samenko2021fine}. Whenever a token is in ${\cal V}_{in}$ but not in ${\cal V}_{gen}$, VIPI calculates all the partitions of the new token with tokens from ${\cal V}_{gen}$, then takes the minimal partitions and finally averages them to obtain an embedding for the new token. 
Differently, we define a simplified implementation of VIPI called FVT for Fast Vocabulary Transfer. Instead of calculating all tokenizations, FVT uses a straightforward assignment mechanism, whereby each token $t_i \in {\cal V}_{in}$ is partitioned using $\tokgen$. If $t_i$ belongs to both vocabularies,  $t_i \in {\cal V}_{in} \cap {\cal V}_{gen}$, then $\tokgen{(t_i)}=t_i$ and the in-domain LM embedding $E_{in}(t_i)$ is the same as the embedding in the general LM: 
\begin{equation}
E_{in}(t_i) = E_{gen}(t_i).\label{eq:base_vt}    
\end{equation}
If instead  $t_i \in {\cal V}_{in} \setminus {\cal V}_{gen}$, then the in-domain embedding is the average of the embeddings associated with the tokens produced by $\tokgen$: 
 
\begin{equation}
    E_{in}(t_i) = \frac{1}{{|\tokgen(t_i)|}} \cdot \sum_{t_j \in \tokgen(t_i) } E_{gen}(t_j). \label{eq:fvt}
\end{equation}
Please notice that Equation~\ref{eq:fvt} is a generalization of Equation~\ref{eq:base_vt}. Indeed, in case $t_i \in {\cal V}_{in} \cap  {\cal V}_{gen}$, Equation~\ref{eq:fvt} falls back to Equation~\ref{eq:base_vt}.

Once embeddings are initialized with FVT, we adjust the model's weights by training it with MLM on the in-domain data before fine-tuning it on the downstream task. MLM eases adaptation and has already been found to be beneficial in~\cite{samenko2021fine}. We observed this trend as well during preliminary experiments, therefore we kept such a tuning stage in all our experiments.

\begin{figure}
\noindent\fbox{\begin{minipage}{19.5em}
\footnotesize
\textbf{Input:}
\\
\texttt{ He was initially treated with interferon alfa.}
\\ 
\\
\textbf{$\tokgen$:}\\ \texttt{ He, was, initially, treated, with, \textbf{inter},\textbf{\#\#fer},\textbf{ \#\#on}, \textbf{al}, \textbf{\#\#fa}, .
\\
\\
\textbf{$\tok_{100}$:}\\ \texttt{He, was, initially, treated, with, \textbf{interferon}, \textbf{alfa}, .}
}
\end{minipage}}\caption{Example of different tokenizations using a pre-trained or an adapted tokenizer. In the latter case, domain-specific words are not broken down into multiple word pieces.}\label{fig:vt_examples}
\end{figure}

\begin{table}[]
    \centering
    \begin{tabular}{cccccc}
    \hline
    \textbf{Dataset} & $\tokgen$ & $\tok_{100}$ & $\tok_{75}$ & $\tok_{50}$ & $\tok_{25}$\\ \hline
    \textbf{ADE} & 31 & 21	& 22 & 23 & 26\\
    \textbf{LEDGAR} & 155 & 131 & 131 & 132 & 135\\
    \textbf{CoNLL03} & 19 & 17 & 17 & 18 & 20\\ \hline
    \end{tabular}
    \caption{Average sequence length on the three datasets with different tokenizers. $\tokgen$ is the generic tokenizer (BERT cased), the same in each corpus, while $\mathcal{T}_{\%}$  are the tokenizers trained in the vertical domain itself, where ${\%}$ indicates the percentage of the original vocabulary size that has been set for training it.}
    \label{tab:seq_len}
\end{table}



As a baseline model, we also implement a method called Partial Vocabulary Transfer (PVT), whereby only the tokens belonging to both vocabularies $t_i \in {\cal V}_{in} \cap  {\cal V}_{gen}$ are initialized with pre-trained embeddings, while unseen new tokens are randomly initialized.

\begin{table}[ht]
    \begin{tabular}{cccc}
        \hline
        
        \textbf{Transfer} & \textbf{ADE} & \textbf{LEDGAR} & \textbf{CoNLL03}\\ \hline
        
        $\tokgen$ & \textbf{90.80} & 80.93 & \textbf{89.43}\\ \hline
        
        $\tok_{100}$ + FVT & 90.77 & 80.60          & 87.87\\
        $\tok_{75}$  + FVT & 90.40 & 80.93          & 87.90\\
        $\tok_{50}$  + FVT & 90.07 & 80.93          & 86.87\\
        $\tok_{25}$  + FVT & 90.27 & \textbf{81.03} & 86.17\\ \hline
        
        $\tok_{100}$ + PVT & 82.57 & 80.07 & 84.53\\
        $\tok_{75}$  + PVT & 82.47 & 80.33 & 84.63\\
        $\tok_{50}$  + PVT & 83.07 & 80.23 & 84.43\\
        $\tok_{25}$  + PVT & 83.57 & 80.20 & 83.47\\ 
        
        \hline             
    \end{tabular} \caption{F1 results on the three benchmarks. A pre-trained language model fine-tuned on the task ($\tokgen$) is compared with models having differently sized in-domain tokenizers ($\tok_{100},\tok_{75},\tok_{50},\tok_{25}$) adapted by transferring information with FVT or PVT.}\label{tab:base_f1_results}
\end{table}

  
  
  



\subsection{Distillation}
VT can be combined with other model compression methods like quantization, pruning and KD. For some of the methods, the combination is trivial, since they have no impact on the vocabulary. KD, however, requires the vocabularies of the student and teacher to be aligned. Hence, its integration with VT is non-trivial. Accordingly, we set up a KD procedure with VT, in order
to determine the effects of applying both VT and KD to an LM. 

Our distillation consists of two steps. In the first step, we replicate the distillation process used in \cite{distilbert} for DistilBERT, in which the number of layers of the encoder is halved and a triple loss-function is applied: a distillation loss, a MLM loss, and a cosine embedding loss. However, unlike the original setup, we do not remove the token-type embeddings and pooler.
Inspired by  \citet{gordon-duh-2020-distill}, after distilling the student on ${\cal D}_{gen}$, we further distil the student using  ${\cal D}_{in}$. 
However, instead of adapting the teacher before the second distillation, we simply distil the student a second time on the in-domain dataset. Finally, we apply VT using either FVT or PVT and fine-tune the student model on the in-domain datasets.

Our choice of applying VT after KD is based on findings by~\citet{kim-hassan-2020-fastformers}, that different input embedding spaces will produce different output embedding spaces. This difference in spaces is not conducive to knowledge transfer during distillation. Hence, if VT were to be applied first to the student, its input embedding space would differ greatly from that of the pre-trained teacher during distillation.

\section{Experiments}\label{sec:experiments}
In the experiments we measure the impact of FVT on three main KPIs: quality (F1 score), size of the models and speedup in inference.

\begin{figure*}[h]
  \centering
  \includegraphics[width=.325\textwidth]{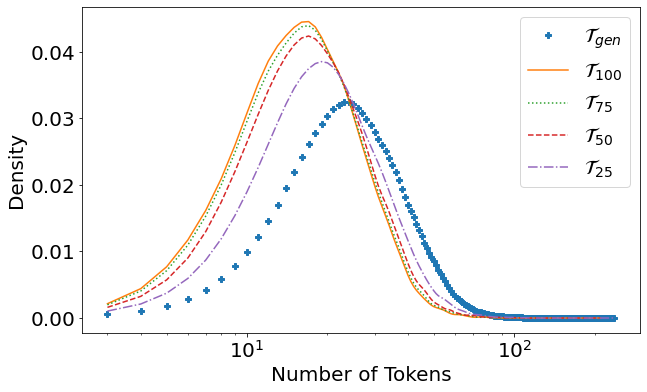} 
  \includegraphics[width=.332\textwidth]{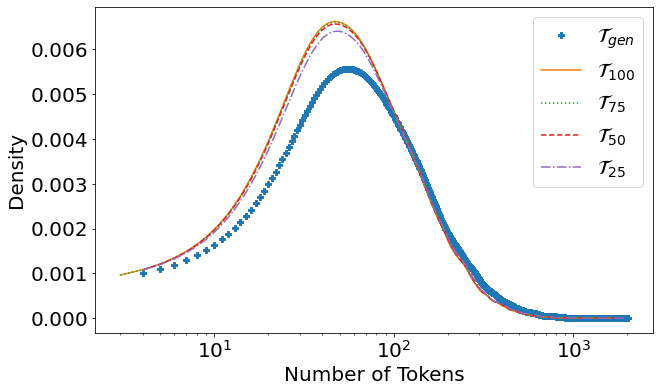}
  \includegraphics[width=.325\textwidth]{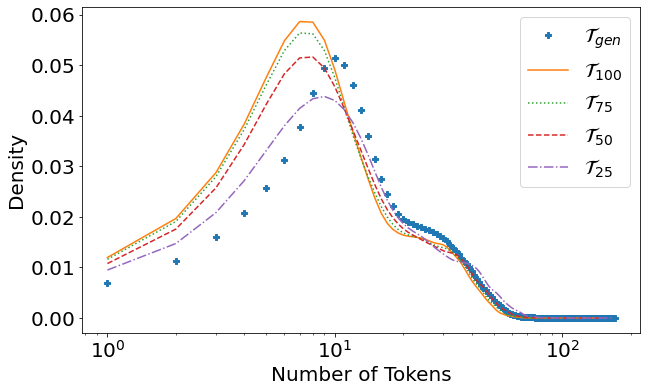}
  \caption{ Sequence length distribution of each tokenizer on ADE, LEDGAR and CoNLL03 (left to right).}\label{fig:tok_distribution}
\end{figure*}

\begin{table}[]
    \begin{tabular}{cccc}
        \hline
        
        \multicolumn{4}{c}{\textbf{Distillation}}\\
        
        \textbf{Transfer} & \textbf{ADE} & \textbf{LEDGAR} & \textbf{CoNLL03}\\ \hline

        $\tokgen$ & \textbf{90.47} & 78.37 & \textbf{86.90}\\ \hline
        
        $\tok_{100}$ + FVT & 89.47 & 78.33          & 84.63\\
        $\tok_{75}$  + FVT & 88.57 & 78.90          & 84.23\\
        $\tok_{50}$  + FVT & 88.43 & \textbf{79.30} & 83.80\\
        $\tok_{25}$  + FVT & 88.23 & 78.10          & 83.13\\ \hline

        $\tok_{100}$ + PVT & 79.13 & 76.97 & 81.13\\
        $\tok_{75}$  + PVT & 78.87 & 76.93 & 81.40\\
        $\tok_{50}$  + PVT & 76.30 & 77.37 & 81.63\\
        $\tok_{25}$  + PVT & 77.90 & 77.33 & 79.50\\ 
        
        \hline             
    \end{tabular} \caption{F1 results on the three benchmarks. A distilled language model fine-tuned on the task ($\tokgen$) is compared with models having differently sized in-domain tokenizers ($\tok_{100},\tok_{75},\tok_{50},\tok_{25}$) adapted by transferring information with FVT or PVT.}\label{tab:distill_f1_results}
\end{table}

\subsection{Experimental Setup}
We consider for all our experiments the pre-trained cased version of $\text{BERT}_{base}$~\cite{bert} as our pre-trained language model. Its tokenizer is composed of 28996 wordpieces.
We then define four vocabulary sizes for retraining our tokenizers. Specifically, we take the original vocabulary size and define it as a vocabulary size of 100\%. We subsequently reduce this size to 75\%, 50\%, and 25\%. From now on, we will refer to such tokenizers as $\tok_{100},\tok_{75},\tok_{50},\tok_{25}$ respectively, while the original vocabulary will be called $\tokgen$.

Models are fine-tuned for 10 epochs with early stopping on the downstream task. We set the initial learning rate to $3\cdot10^{-5}$ and batch size to 64 for each task. The sequence length is set to 64 for ADE and CoNLL03 and 128 for LEDGAR. Each configuration is repeated 3 times with different random initializations. MLM is performed for one epoch.

\subsection{Datasets}
To best assess the effectiveness of VT, we apply it on three different tasks from three heterogeneous linguistic domains: medical (ADE), legal (LEDGAR) and news (CoNLL03). 
Table~\ref{tab:data_stats} reports the dataset statistics.

  
  \paragraph{ADE.} The Adverse Drug Events (ADE) corpus \citep{GURULINGAPPA2012885} is a binary sentence classification dataset in the medical domain. This domain is particularly suitable for investigating the benefits of VT, since documents are characterized by the presence of frequent technical terms, such as drug and disease names, that are usually rare in common language. Domain-specific words are usually split into multiple tokens, yielding longer sequences and breaking the semantics of a word into multiple pieces. An example is shown in Figure~\ref{fig:vt_examples}. 

  \paragraph{LEDGAR.} LEDGAR~\cite{tuggener2020ledgar} is a document classification corpus of legal provisions in contracts from the US Securities and Exchange Commission (SEC). The dataset is annotated  with 100 different mutually-exclusive labels. It is also part of LexGLUE~\cite{chalkidis-etal-2022-lexglue}, a benchmark for legal language understanding.

  \paragraph{CoNLL03.} CoNLL03~\cite{tjong-kim-sang-de-meulder-2003-introduction} is a popular Named Entity Recognition (NER) benchmark. It is made of news stories from the Reuters corpus. We chose this corpus because, differently from ADE and LEDGAR, the news domain typically uses a more standard language, hence we expect its distribution to differ less from the one captured by a general-purpose tokenizers in the web. Statistics in Table~\ref{tab:seq_len} confirms this hypothesis. We can observe that the sequence compression gain obtained with domain-specific tokenizers is less significant with respect to LEDGAR and ADE.
    
\begin{table}[h]
\centering
    \begin{tabular}{cccc}
        \hline
        \textbf{Dataset} & \textbf{Train} & \textbf{Validation} & \textbf{Test}\\ \hline
        
        ADE & 16716 & 3344 & 836\\
        LEDGAR & 60000 & 10000 & 10000\\
        CoNLL03 & 14042 & 3251 & 3454\\
        
        \hline
    \end{tabular}
    \caption{Number of examples of each dataset.}
    \label{tab:data_stats}
\end{table}

\begin{table*}[ht]
    \setlength\tabcolsep{4.1pt}
    \begin{tabular}{cccccccccc}
        \hline
        \multirow{2}{*}{\textbf{Transfer}} & 
        \multicolumn{3}{c}{\textbf{ADE}} & \multicolumn{3}{c}{\textbf{LEDGAR}} & \multicolumn{3}{c}{\textbf{CoNLL03}} \\
        
         & $\Delta$\textbf{F1} & $\Delta$\textbf{Size} & \textbf{Speedup} & 
         $\Delta$\textbf{F1} & $\Delta$\textbf{Size} & \textbf{Speedup} &
         $\Delta$\textbf{F1} & $\Delta$\textbf{Size} & \textbf{Speedup}  \\ \hline
        
        $\tokgen$ & 90.80 & 433.32 & 1.00 & 80.93 & 433.62 & 1.00 & 89.43 & 430.98 & 1.00\\ \hline

        $\tok_{100}$ + FVT & \textbf{-0.04} & 0.00   & 1.40 & -0.41         & 0.00   & 1.21 & -1.75          & 0.00   & 1.07\\
        $\tok_{75}$  + FVT & -0.44          & -5.14  & 1.35 & 0.00          & -5.14  & 1.21 & \textbf{-1.71} & -5.17  & 1.07\\
        $\tok_{50}$  + FVT & -0.81          & -10.28 & 1.32 & 0.00          & -10.27 & 1.10 & -2.87          & -10.33 & 1.02\\
        $\tok_{25}$  + FVT & -0.59          & -15.42 & 1.20 & \textbf{0.12} & -15.41 & 1.09 & -3.65          & -15.50 & 0.99\\ \hline
        
        Distil + $\tok_{100}$ + FVT & -1.47 & -39.26 & \textbf{2.76} & -3.21 & -39.24 & \textbf{2.38} & -5.37 & -39.48 & \textbf{2.11}\\
        Distil + $\tok_{75}$  + FVT & -2.46 & -44.40 & 2.64 & -2.51 & -44.37 & \textbf{2.38} & -5.81 & -44.64 & \textbf{2.11}\\
        Distil + $\tok_{50}$  + FVT & -2.61 & -49.54 & 2.59 & -2.02 & -49.51 & 2.16 & -6.30 & -49.81 & 2.01\\
        Distil + $\tok_{25}$  + FVT & -2.83 & \textbf{-54.68} & 2.37 & -3.50 & \textbf{-54.64} & 2.14 & -7.04 & \textbf{-54.98} & 1.96\\ \hline
    \end{tabular} \caption{The first row ($\tokgen$) reports absolute values of the LM fine-tuned on the downstream task without VT or KD. The rows below show values relative to $\tokgen$.}\label{tab:relative results}
\end{table*}

\begin{figure}[ht]
  \centering
  \includegraphics[width=\columnwidth]{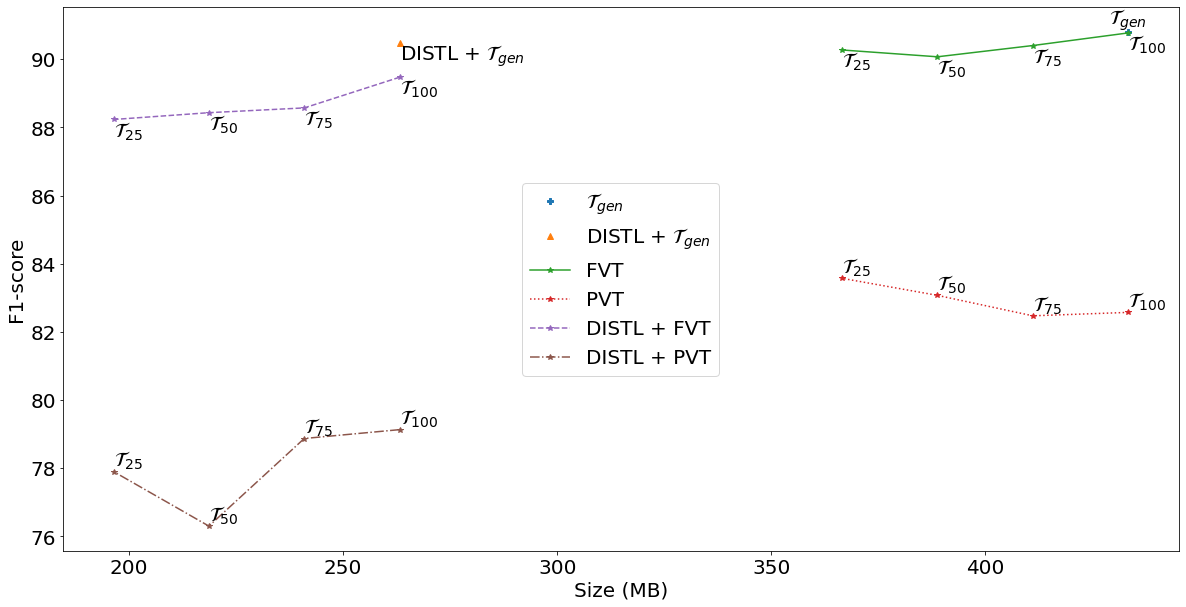} 
  \caption{F1-score vs model size of VT with or without KD on ADE. VT and KD together can further compress a LM's size in exchange for a limited performance drop. FVT is better than PVT. A smaller vocabulary size does not always imply a lower performance.}\label{fig:2d_size_f1f1}
\end{figure}

\subsection{Results}
We report an extensive evaluation of FVT on different setups and perspectives.

\paragraph{In-domain Tokenization.}
By retraining the tokenizer on the in-domain dataset, the average number of tokens per sequence decreases since the learned distribution reduces the number of word splits, as shown in Table~\ref{tab:seq_len}. In the medical domain, which is particularly specialized, we notice a remarkable 32\% reduction of the average number of tokens per sequence. We expect this to yield a noticeable impact on inference time speedup. Furthermore, we can notice in Figure~\ref{fig:tok_distribution} that the sequence length distribution shifts to the left for the learned tokenizers. It can also be observed that by reducing the vocabulary size of the \textit{in-domain} tokenizer, the sequence length distribution will begin to shift back to the right. Indeed, with fewer tokens in its vocabulary, the tokenizer will need to break down words more frequently into subwords.

\paragraph{Vocabulary Transfer.}
From the results shown in Tables~\ref{tab:base_f1_results} and \ref{tab:distill_f1_results}, we note a few interesting findings. First, FVT vectors initialization method consistently outperforms the baseline PVT, which confirms the positive contribution of Equation~\ref{eq:fvt}. Second, transferring vocabulary with FVT causes limited drops in performance, especially in LEDGAR (the largest one), where F1 slightly increases despite a 75\% vocabulary reduction. As a matter of fact, the effects of FVT on model performance do not have a steadily decreasing trend as might be expected when reducing the vocabulary size, as also evident from Figure~\ref{fig:2d_size_f1f1}. In some cases, somewhat surprisingly, reducing the vocabulary size yields better model performance. 
In other cases, a 50\% vocabulary size reduction yields better results than a full scale reduction or no reduction. Hence, vocabulary size should be considered as a hyperparameter.

\paragraph{Vocabulary Transfer and Distillation.} 
The results summarized in Table~\ref{tab:distill_f1_results} clearly indicate that KD is complementary to VT: there is no harm in applying them together, in terms of performance on the downstream task. Crucially, this guarantees a full exploitation of FVT in the scope of language model compression.

\paragraph{Compression and Efficiency.} After showcasing that VT has limited impact on performance, we analyze and discuss its effects on efficiency and model compression. Table~\ref{tab:relative results} reports the relative F1 drop on the downstream task with respect to the original LM ($\Delta$F1), the relative reduction in model size ($\Delta$Size) and the speedup gained by FVT alone and by FVT combined with KD for varying vocabulary sizes. 
Either way, FVT achieves a remarkable 15\%+ reduction with respect to BERT's learnable parameters, with almost no loss in F1. 

Furthermore, the reduced input length enabled by in-domain tokenization brings a reduction in inference time. The more a language is specialized, the higher is the speedup with in-domain tokenizers. This is also confirmed by the experiments, where the major benefits are obtained on the medical domain, with a x1.40 speedup. In CoNLL03 instead where language is much less specialized, speedup reduces and even disappears with $\tok_{25}$. 
Distillation further pushes compression and speedup in any benchmark and setup, up to about 55\% (of which 15\% due to VT) and x2.75 respectively. 

In summary, depending on the application needs, VT enables a strategic trade-off between compression rate, inference speed and accuracy.

\section{Conclusion}\label{sec:conclusions}
The viability and success of industrial NLP applications often hinges on a delicate trade-off between computational requirements, responsiveness and output quality. Hence, language model compression methods are an active area of research whose practical ramifications are self-evident.  One of the factors that greatly contribute to a model's inference speed and memory footprint is vocabulary size. VT has been recently proposed for improving performance, but never so far in the scope of model compression. In this work, we run an extensive experimental study on the application of a lightweight method for VT, called FVT. An analysis conducted on various downstream tasks, application domains, vocabulary sizes and on its possible combination with knowledge distillation indicates that FVT enables a strategic trade-off between compression rate, inference speed and accuracy, especially, but not only, in more specialized domains. Importantly, FVT appears to be orthogonal to other model compression methods.

In the future, we plan to fully integrate Vocabulary Transfer within Knowledge Distillation during the learning process in order to maximize the information transfer.


\bibliographystyle{acl_natbib}
\bibliography{anthology,custom}

\end{document}